\documentclass[letterpaper, 10 pt, conference]{ieeeconf}  

\IEEEoverridecommandlockouts                              

\usepackage{amsmath,amssymb,amsfonts}
\usepackage{cite}
\usepackage{graphicx}
\usepackage{textcomp}
\usepackage{xcolor}
\usepackage{booktabs}
\usepackage{hyperref}
\usepackage{balance}

\usepackage{siunitx}

\usepackage{tikz}
\usetikzlibrary{arrows.meta, patterns, calc}
\usepackage{pgfplots}
\pgfplotsset{compat=1.17}

\hypersetup{
    colorlinks=true,
    linkcolor=blue,
    citecolor=blue,
    urlcolor=blue
}

\newif\ifsectionbreaks
\newcommand{\sectionbreak}{\ifsectionbreaks\clearpage\fi}

\newcommand{\vc}{\mathbf{c}}            
\newcommand{\cbar}{\bar{\mathbf{c}}}    
\newcommand{\cx}{c_x}
\newcommand{\dcx}{\dot{c}_x}     
\newcommand{\ddcx}{\ddot{c}_x}   
\newcommand{\cxbar}{\bar{c}_x}   
\newcommand{\cy}{c_y}
\newcommand{\vF}{\mathbf{F}}            
\newcommand{\Fbar}{\bar{\mathbf{F}}}    
\newcommand{\vq}{\mathbf{q}}            
\newcommand{\vtau}{\boldsymbol{\tau}}   
\newcommand{\vp}{\mathbf{p}}            
\newcommand{\va}{\mathbf{a}}            
\newcommand{\vt}{\mathbf{t}}            
\newcommand{\vn}{\mathbf{n}}            
\newcommand{\vg}{\mathbf{g}}            
\newcommand{\gcomp}[2]{g_{#1,#2}}      
\newcommand{\mJ}{\mathbf{J}}            
\newcommand{\mM}{\mathbf{M}}            
\newcommand{\mC}{\mathbf{C}}            
\newcommand{\mD}{\mathbf{D}}            
\newcommand{\mK}{\mathbf{K}}            
\newcommand{\mH}{\mathbf{H}}            
\newcommand{\mI}{\mathbf{I}}            
\newcommand{\qref}{q_\mathrm{ref}}      
\newcommand{\kp}{k_\mathrm{p}}          
\newcommand{\kd}{k_\mathrm{d}}          
\newcommand{\If}{I_\mathrm{f}}          
\newcommand{\xtip}{c_{x,\mathrm{tip}}}  
\newcommand{\estar}{e^\ast}             

\title{\LARGE \bf
    Input-output characterization of robotic grasping:  
    \\ Connecting design parameters to grasp behavior through the admittance
}

\author{
    Julien Vanderheyden$^1$, 
    Guillaume Drion$^1$,
    Pierre Sacré$^{1}$, 
    Fulvio Forni$^{2}$ 
    \thanks{$^{1}$J.\ Vanderheyden, G.\ Drion, and P.\ Sacré are with the Department of Electrical Engineering and Computer Science, University of Liège, Belgium (julien.vanderheyden@uliege.be; gdrion@uliege.be; p.sacre@uliege.be).}
    \thanks{$^{2}$F.\ Forni is with the Department of Engineering, University of Cambridge, United Kingdom (f.forni@eng.cam.ac.uk).} 
}

\usepackage{etoolbox}
\newlength{\topfloatdrop}
\makeatletter
\patchcmd{\@cflt}{\unvbox\@tempboxa}{\vskip\topfloatdrop\unvbox\@tempboxa}{}{\typeout{PATCH cflt 1 FAILED}}
\patchcmd{\@cflt}{\vskip \textfloatsep}{\vskip \glueexpr\textfloatsep-\topfloatdrop\relax}{}{\typeout{PATCH cflt 2 FAILED}}
\patchcmd{\@cflb}{\vskip \textfloatsep}{\vskip \glueexpr\textfloatsep-\topfloatdrop\relax}{}{\typeout{PATCH cflb FAILED}}
\makeatother

\usepackage{xcolor}

\definecolor{lightgrey}{RGB}{159,159,159}      
\definecolor{verylightgrey}{RGB}{231,231,231}  

\begin{document}

\maketitle
\thispagestyle{empty}
\pagestyle{empty}

\begin{abstract}
Low-impedance grippers make the dynamics of a grasp matter. Current grasp metrics, however, are static: they describe how a held object resists constant loads, not how it responds to time-varying ones. To quantify this, 
we rely on small-signal analysis to characterize a grasp by its admittance, the transfer matrix relating load perturbations to object displacements around an equilibrium. This transfer matrix recovers the stiffness and stability of the grasp at zero frequency and captures its natural frequencies and peak gain under time-varying loads.
We apply this characterization to a planar two-finger gripper under proportional-derivative control. We show that two grasps of the same object in the same position and with the same contact forces can differ in stiffness by two orders of magnitude. Likewise, two grasps that share the same stiffness can exhibit very different resonances. Nonlinear simulations confirm the effectiveness of our small-signal analysis in capturing the dynamic behavior of grasping.%
\end{abstract}


\section{Introduction}

Grasping with low-impedance grippers is a dynamic problem. Compliance has long been recognized as useful in manipulation~\cite{mason1981compliance, hogan1985impedance, cutkosky1989computing}, but only recently have grippers started to implement it,
thanks to advances in actuator design~\cite{wensing2017proprioceptive}, low-inertia fingers~\cite{lee2026ultra}, direct-drive actuation~\cite{bhatia2019direct}, and elastic elements~\cite{catalano2014adaptive, friedl2020clash}. These grippers promote force transparency, high-bandwidth feedback, and fast, contact-rich motions, and thus challenge the long-standing treatment of grasping and manipulation as quasi-static problems~\cite{roa2015grasp}. Their low stiffness and friction also yield lightly damped resonances at low frequencies, which these motions can excite. Wimb\"ock et al.~\cite{wimbock2012comparison} showed experimentally that the dynamics of such grippers should not be neglected. 

Current grasp metrics, however, do not describe how a held object responds to time-varying loads. Standard criteria, such as the Ferrari--Canny $\epsilon$-metric~\cite{ferrari1992planning} and stiffness-based quality measures~\cite{lin2000stiffness}, evaluate whether a hand can withstand arbitrary wrench disturbance magnitudes in static conditions. They treat finger links as massless and so neglect mechanical inertia, joint damping, and coupling resonances. Yet low-impedance grippers are designed for unstructured environments, where fast inertial perturbations and contact impacts are expected during standard manipulation, such as pushing or insertion. Dynamic metrics exist but address other questions, such as the accelerations a hand can impart to the object~\cite{yokokohji2009dynamic} or the outcome of grasp acquisition under pose uncertainty~\cite{kim2013physically}. A tool is therefore needed to quantify the response of a held object to time-varying loads, and to relate it to the gripper parameters, which include both the morphology parameters and the control parameters.

To quantify this response, we characterized a grasp by its admittance, $\mH(s) = (\mM s^2 + \mD s + \mK)^{-1}$, 
the transfer matrix relating load perturbations to displacements of the grasped object around an equilibrium. In contrast to classical stiffness-based measures that relate static load perturbations to static displacements, the admittance includes inertia and damping, thereby capturing the full spectrum of time-varying perturbations. At zero frequency, the admittance recovers the stiffness; at nonzero frequencies, it reveals critical dynamic behavior, such as resonances, that quasi-static metrics miss. Crucially, the gripper parameters directly shape the admittance, establishing a connection between these design parameters and the grasp dynamics.

\begin{figure}[tbp]
  \centering
\providecommand{\cx}{c_x}
\providecommand{\cy}{c_y}
\definecolor{ctrl}{RGB}{0,0,255}    
\definecolor{load}{RGB}{255,0,0}    
\providecolor{lightgrey}{RGB}{159,159,159}      
\providecolor{verylightgrey}{RGB}{231,231,231}  
\begin{tikzpicture}[scale=0.62, >={Stealth[length=5pt]},
    finger/.style={line width=1.4pt, line cap=round},
    force/.style={->, thick, ctrl},
    dim/.style={|<->|, thin, >={Stealth[length=3.5pt]}, lightgrey},
    ext/.style={thin, lightgrey},
    ann/.style={lab, text=lightgrey},
    lab/.style={font=\small}]

  \def\gripA{4} \def\gripR{3} \def\gripC{0.96} \def\gripS{0.28}
  \coordinate (O)  at (0,0);
  \coordinate (P1) at (0,\gripA);
  \coordinate (P2) at (0,-\gripA);
  \coordinate (X)  at (\gripR,0);                  
  \coordinate (C1) at (3.84, 2.88);                
  \coordinate (C2) at (3.84,-2.88);
  \coordinate (T1) at (9.6, 1.2);                  
  \coordinate (T2) at (9.6,-1.2);

  \fill[verylightgrey] (X) circle (\gripR);   

  \fill[pattern=north east lines, pattern color=gray] (-0.3,-4.4) rectangle (0,4.4);
  \draw[line width=1.4pt] (0,-4.4) -- (0,4.4);
  \node[lab, rotate=90] at (-0.75,-2.4) {palm};

  \fill[load] (X) circle (2.6pt);
  \node[lab, text=load, anchor=north west] at ($(X)+(0.05,-0.12)$) {$\mathbf{c}=(\cx,\cy)$};
  \draw[thin, lightgrey] (X) -- ++(215:\gripR) node[midway, above left=-1pt, ann] {$r$};

  \draw[finger] (P1) -- (T1) node[right, lab] {};
  \draw[finger] (P2) -- (T2) node[right, lab] {};
  \foreach \P in {P1,P2} \draw[thick, fill=white] (\P) circle (3pt);
  \node[lab, anchor=south east] at ($(P1)+(-0.4,0.15)$) {$\mathbf{p}_1=(0,d)$};
  \node[lab, anchor=north east] at ($(P2)+(-0.4,-0.15)$) {$\mathbf{p}_2=(0,-d)$};

  \draw[dim] (-1.35,0) -- (-1.35,\gripA) node[midway, left, ann] {$d$};
  \draw[ext] (-1.55,0) -- (-0.35,0);

  \draw[->, thin, lightgrey] ($(P1)+(1.6,0)$) arc[start angle=0, end angle=-16.26, radius=1.6];
  \node[ann] at ($(P1)+(-7:2.05)$) {$q_1$};
  \draw[->, thin, lightgrey] ($(P2)+(1.6,0)$) arc[start angle=0, end angle=16.26, radius=1.6];
  \node[ann] at ($(P2)+(7:2.05)$) {$q_2$};

  \draw[thin, dashed, lightgrey] (P1) -- ++(3.0,0);   
  \draw[blue!50, thick, dash pattern=on 4pt off 2pt] (P1) -- ++(-30:3.7);
  \draw[->, blue!50, semithick] ($(P1)+(2.75,0)$) arc[start angle=0, end angle=-30, radius=2.75];
  \node[anchor=north east, inner sep=1pt, lab, text=blue!50] at ($(P1)+(-35:2.75)$) {$q_\mathrm{ref}$};
  \draw[thin, dashed, lightgrey] (P2) -- ++(3.0,0);
  \draw[blue!50, thick, dash pattern=on 4pt off 2pt] (P2) -- ++(30:3.7);
  \draw[->, blue!50, semithick] ($(P2)+(2.75,0)$) arc[start angle=0, end angle=30, radius=2.75];
  \node[anchor=south east, inner sep=1pt, lab, text=blue!50] at ($(P2)+(35:2.75)$) {$q_\mathrm{ref}$};

  \foreach \C in {C1,C2} \fill (\C) circle (1.6pt);
  \draw[force] (C1) -- ++(-1.5*\gripS,-1.5*\gripC) node[left, lab, text=ctrl] {$f_1\mathbf{n}_1$};
  \draw[force] (C2) -- ++(-1.5*\gripS, 1.5*\gripC) node[left, lab, text=ctrl] {$f_2\mathbf{n}_2$};

  \node[lab] at ($(C1)+(-61:0.62)$) {$\mathbf{a}_1$};
  \node[lab] at ($(C2)+(61:0.62)$) {$\mathbf{a}_2$};

  \draw[force] (O) -- ++(1.3,0) node[above, lab, text=ctrl] {$f_0\mathbf{n}_0$};
  \draw[->, line width=1.1pt, load] (X) -- ++(4.4,0) node[above left, lab, text=load] {$\mathbf{F}$};

  \draw[ext] (C1) -- ++(0.28*0.85, 0.96*0.85);
  \draw[ext] (P1) -- ++(0.28*0.85, 0.96*0.85);
  \draw[dim] ($(P1)+(0.28*0.65, 0.96*0.65)$) -- ($(C1)+(0.28*0.65, 0.96*0.65)$)
        node[midway, above right=-2pt, ann] {$\ell_1$};

  \draw[ext] (C2) -- ++(0.28*0.85,-0.96*0.85);
  \draw[ext] (T2) -- ++(0.28*1.55,-0.96*1.55);
  \draw[ext] (P2) -- ++(0.28*1.55,-0.96*1.55);
  \draw[dim] ($(P2)+(0.28*0.65,-0.96*0.65)$) -- ($(C2)+(0.28*0.65,-0.96*0.65)$)
        node[midway, below right=-2pt, ann] {$\ell_2$};
  \draw[dim] ($(P2)+(0.28*1.35,-0.96*1.35)$) -- ($(T2)+(0.28*1.35,-0.96*1.35)$)
        node[midway, below right=-2pt, ann] {$L$};

  \begin{scope}[shift={(8.6,-5.2)}]
    \draw[->, thin, lightgrey] (0,0) -- (0.9,0) node[right, ann] {$x$};
    \draw[->, thin, lightgrey] (0,0) -- (0,0.9) node[above, ann] {$y$};
  \end{scope}
\end{tikzpicture}%
  \caption{
  Planar gripper model: the palm and the two fingers push on the disk along their normals, with the forces~$f_0\vn_0$ and $f_i\vn_i$, to balance the load $\vF = (F_x, F_y)$ acting on the disk center~$\vc$. Finger~$i$ rotates about $\vp_i$, and its angle~$q_i$ is measured from the $x$ direction, positive when closing. It touches the disk at $\va_i$, at distance~$\ell_i$ from its joint. A PD controller drives each finger toward the reference~$\qref$. The disk is drawn on the palm.}
  \label{fig:model}
\end{figure}
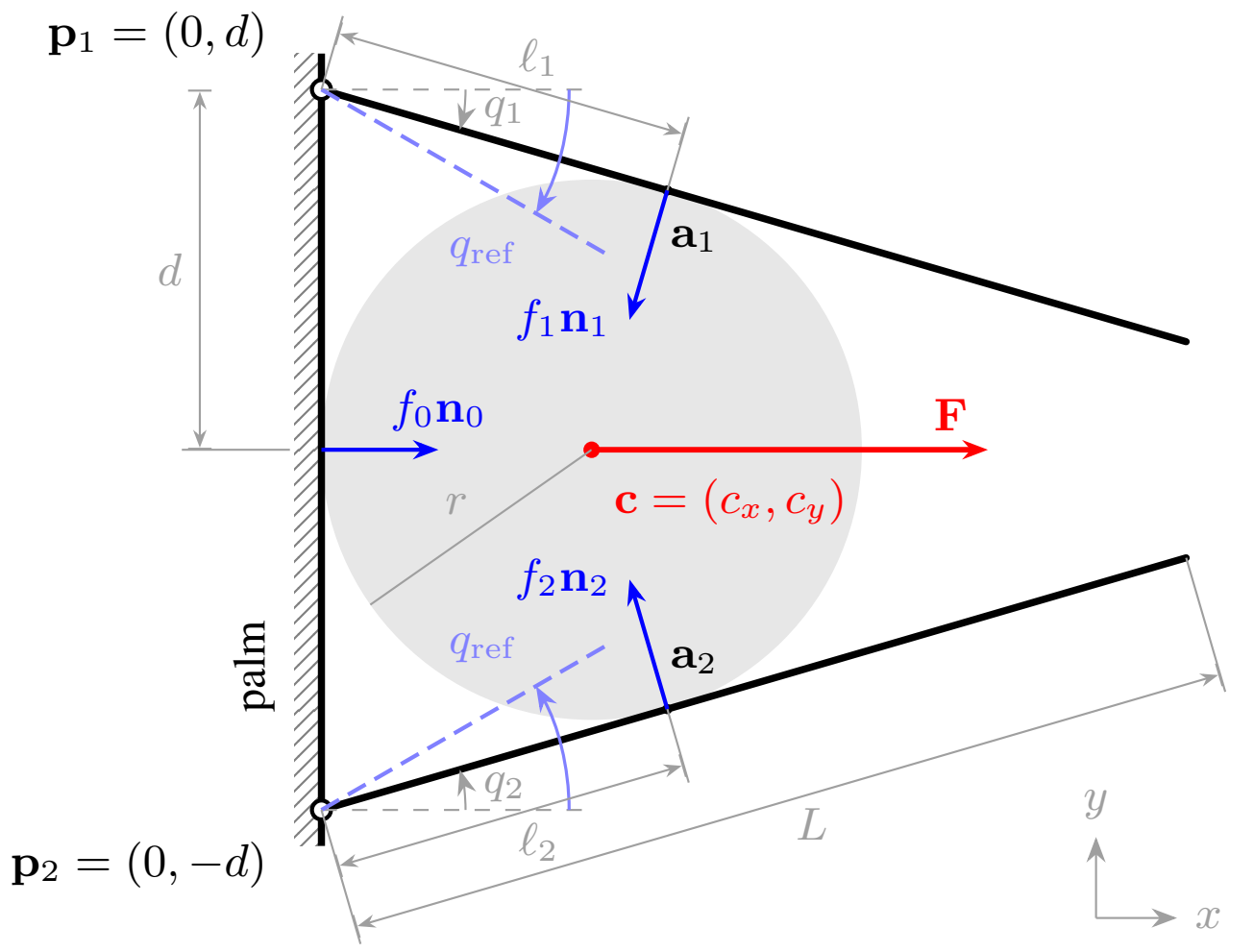

On a minimal gripper (Fig.~\ref{fig:model}), we showed that grasps that are equivalent under static metrics can respond differently to fast load perturbations. Closed-form expressions are hard to derive in general without dropping terms that can destabilize the grasp~\cite{prattichizzo1998dynamic}. We therefore considered a minimal model that keeps these terms and remains analytically tractable: a planar two-finger gripper under joint proportional-derivative (PD) control, holding a disk through rigid frictionless contacts. With this model, we also showed that the same gripper can hold an object in qualitatively different ways. Nonlinear simulations confirmed that the admittance, although derived from a linearization, predicts the main dynamic features of the nonlinear model.

Section~\ref{sec:model} introduces the model used throughout the paper. 
Section~\ref{sec:linearization} presents the input-output characterization of a grasp by its admittance, obtained by linearization.
Section~\ref{sec:static} then derives the static equilibria and their stability. 
Section~\ref{sec:dynamic} studies the dynamic response, showing how the gripper parameters shape the admittance~$\mH(s)$ and how far this admittance predicts the nonlinear response. 
Finally, Section~\ref{sec:conclusion} discusses the results and their limitations. The main symbols and the parameter values
are summarized in Table~\ref{tab:symbols}.

\begin{table}[tbp]
  \centering
  \caption{Parameters and variables.}
  \label{tab:symbols}
  \footnotesize
  \begin{tabular}{@{}l l l@{}}
    \toprule
    Symbol & Meaning & Value (nominal)\\
    \midrule
    \multicolumn{3}{@{}l}{\emph{Gripper parameters: morphology}} \\
    $L$ & finger length & \qty{100}{\milli\metre} \\
    $d$ & half-distance between the joints & \qty{40}{\milli\metre} \\
    $\If$ & finger inertia about its joint & \qty{6.7e-5}{\kilo\gram\metre\squared} \\
    \multicolumn{3}{@{}l}{\emph{Gripper parameters: control}} \\
    $\kp$ & proportional gain & \qty{1}{\newton\metre\per\radian} \\
    $\kd$ & derivative gain & \qty{0.01}{\newton\metre\second\per\radian} \\
    $\qref$ & joint reference & \qty{20}{\degree}, \qty{90}{\degree} \\
    \multicolumn{3}{@{}l}{\emph{Disk and load}} \\
    $r$, $m$ & disk radius and mass & \qty{30}{\milli\metre}, \qty{100}{\gram} \\
    $\vF = (F_x, F_y)$ & load & (\qty{0.6}{\newton}, \qty{0.0}{\newton})\\
    \midrule
    \multicolumn{3}{@{}l}{\emph{Variables}} \\
    $\vc = (\cx, \cy)$ & \multicolumn{2}{l@{}}{disk center} \\
    $q_i$, $e_i$ & \multicolumn{2}{l@{}}{finger angle, error} \\
    $\tau_i$ & \multicolumn{2}{l@{}}{joint torque} \\
    $\ell_i$, $h_i$ & \multicolumn{2}{l@{}}{contact distance, distance along $y$} \\
    $\va_i$ & \multicolumn{2}{l@{}}{contact point of finger~$i$} \\
    $\vt_i$, $\vn_i$, $\vn_0$ & \multicolumn{2}{l@{}}{finger tangent and normal, palm normal} \\
    $f_i$, $f_0$ & \multicolumn{2}{l@{}}{finger force, palm force} \\
    $\vg_i$, $\mJ$ & \multicolumn{2}{l@{}}{transmission vector, Jacobian} \\
    $V$ & \multicolumn{2}{l@{}}{controller potential} \\
    $\mM$, $\mC$, $\mD$, $\mK$ & \multicolumn{2}{l@{}}{inertia, Coriolis, damping, and stiffness matrices} \\
    $\mH(s)$ & \multicolumn{2}{l@{}}{admittance (load to displacement)} \\
    $P(\cx)$, $\sigma(\cx)$ & \multicolumn{2}{l@{}}{holding force, transmission} \\
    $\estar(\cx)$, $\kappa$ & \multicolumn{2}{l@{}}{critical error, normalized gain} \\
    \bottomrule
  \end{tabular}
\end{table}

\section{Gripper model and control}\label{sec:model}

\subsection{System description and assumptions}
\label{sec:sys}

We consider a minimal model: a planar two-finger gripper holding a disk through rigid frictionless contacts, with a PD controller at each joint, as shown in Fig.~\ref{fig:model}. In control theory, simple examples such as the inverted pendulum teach general principles. This model plays the same role for grasping.
The model captures the contact geometry, the controller, and the inertia. It treats the fingers and the disk as rigid, and the contacts as rigid and frictionless, so that all the stiffness and the damping of the grasp come from the controller. 

\subsubsection*{Gripper morphology} The gripper has two rigid single-link fingers of length~$L$, attached to a fixed palm by revolute joints. The joints are at distance~$d$ from the center of the palm, and each finger has inertia~$\If$ about its joint. The frame origin is the center of the palm, with $x$ pointing away from the palm and $y$ along it. Each finger angle~$q_i$ is measured from the $x$ direction, positive when closing, that is, clockwise for finger~1 and counterclockwise for finger~2.

\subsubsection*{Disk and load} The fingers hold a rigid disk of radius~$r$ and mass~$m$, with center $\vc = (\cx, \cy)$. We assume $d > r$: otherwise the fingers lean outward to touch the disk, and their torque pushes it out of the gripper instead of holding it. An external load $\vF = (F_x, F_y)$ acts on the disk center. Gravity is perpendicular to the plane of motion.

\subsubsection*{Contacts} The contacts are rigid, frictionless, and unilateral. The bodies cannot interpenetrate, so that $\cx \ge r$, the contact forces act along the contact normals, and the fingers and the palm can only push on the disk. Each finger touches the disk on its side, not at its tip. Without friction, the contacts exert no torque on the disk, so its rotation is decoupled and not modeled.

\subsubsection*{Gripper controller} Both fingers track a common constant reference~$\qref$ with independent PD controllers that share the proportional gain $\kp > 0$ and the derivative gain $\kd > 0$. The joint torques are
\begin{equation}
  \tau_i = \kp e_i + \kd \dot{e}_i ,
  \label{eq:pd}
\end{equation}
where $e_i = \qref - q_i$ is the error of finger~$i$ and $\dot e_i = -\dot q_i$.

\subsection{Kinematics}
\label{sec:kin}
\begin{figure}
  \centering
\providecommand{\cx}{c_x}
\providecommand{\cy}{c_y}
\providecolor{load}{RGB}{255,0,0}               
\providecolor{lightgrey}{RGB}{159,159,159}      
\providecolor{verylightgrey}{RGB}{231,231,231}  
\providecolor{legtan}{RGB}{0,114,178}           
\providecolor{legax}{RGB}{213,94,0}             
\begin{tikzpicture}[scale=0.72, >={Stealth[length=4pt]},
    finger/.style={line width=1.4pt, line cap=round},
    unitvec/.style={->, semithick},
    lab/.style={font=\small},
    ann/.style={lab},
    con/.style={thin},
    leg/.style={line width=1.1pt}]

  \coordinate (P1) at (0,4);
  \coordinate (X)  at (3.5,0.5);                   
  \coordinate (C1) at (4.222,2.365);               
  \coordinate (Q1) at (3.5,4);                     

  \fill[verylightgrey] (X) circle (2);   
  \fill[pattern=north east lines, pattern color=gray] (-0.25,-1.6) rectangle (0,4.6);
  \draw[line width=1.4pt] (0,-1.6) -- (0,4.6);

  \draw[finger] (P1) -- ($(P1)+(4.8*0.9325,-4.8*0.3612)$);   
  \draw[thick, fill=white] (P1) circle (3pt);
  \node[lab, anchor=east] at ($(P1)+(-0.2,0)$) {$\mathbf{p}_1$};
  \fill (C1) circle (1.6pt);

  \node[lab, anchor=south west] at (C1) {$\mathbf{a}_1$};

  \draw[finger, legtan, line width=1.6pt, line cap=butt] (P1) -- (C1);
  \node[ann, text=legtan] at ($(P1)!0.62!(C1)+(0.36*0.45,0.93*0.45)$) {$\ell_1$};
  \draw[leg, legtan] (X) -- (C1) node[pos=0.2, right, ann] {$r$};
  \draw[con, legtan] ($(C1)+(-0.3*0.9325,0.3*0.3612)$) -- ++(-0.3*0.3612,-0.3*0.9325)
        -- ($(C1)+(-0.3*0.3612,-0.3*0.9325)$);

  \draw[leg, legax] (P1) -- (Q1) node[midway, above, ann] {$\cx$};
  \draw[leg, legax] (Q1) -- (X) node[pos=0.55, left, ann] {$h_1$};
  \draw[con, legax] ($(Q1)+(-0.3,0)$) |- ($(Q1)+(0,-0.3)$);
  \draw[con, densely dotted] (P1) -- (X);

  \draw[thick, fill=white] (P1) circle (3pt);
  \fill (C1) circle (1.6pt);

  \draw[->, con] ($(P1)+(1.1,0)$) arc[start angle=0, end angle=-21.17, radius=1.1];
  \node[ann] at ($(P1)+(-10:1.55)$) {$q_1$};
  \draw[con] ($(P1)+(-21.17:1.7)$) arc[start angle=-21.17, end angle=-45, radius=1.7];
  \node[ann] at ($(P1)+(-33:2.2)$) {$\psi_1$};

  \draw[unitvec] (C1) -- ++(-1.0*0.3612,-1.0*0.9325) node[right, lab] {$\mathbf{n}_1$};
  \draw[unitvec] (C1) -- ++(1.2*0.9325,-1.2*0.3612) node[above right=-2pt, lab] {$\mathbf{t}_1$};

  \fill[] (X) circle (2.6pt);
  \node[lab, anchor=north west] at ($(X)+(0.05,-0.1)$) {$\mathbf{c}$};
\end{tikzpicture}%
  \caption{
  The finger angle in \eqref{eq:geometry} follows from two right triangles that share the dotted hypotenuse from $\vp_1$ to $\vc$ (finger~1, disk off the axis). 
  The tangency triangle (\emph{blue}) has legs~$\ell_1$ along $\vt_1$ and $r$ along $\vn_1$, with the right angle at the contact point~$\va_1$. The other triangle (\emph{orange}) has legs~$\cx$ and $h_1 = d - \cy$. 
  The finger angle is the angle of the hypotenuse, $\arctan(h_1/\cx)$, minus $\psi_1 = \arctan(r/\ell_1)$.}
  \label{fig:triangle}
\end{figure}
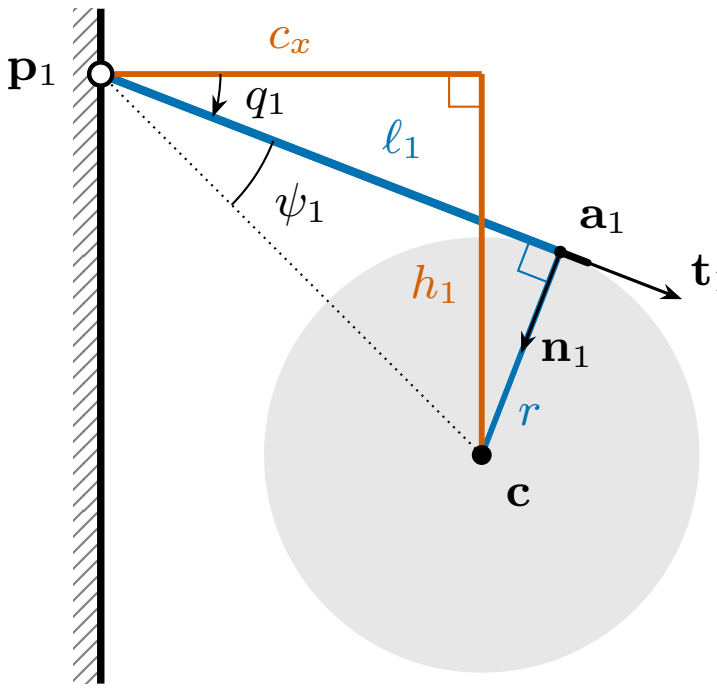

We derive the relationship between the finger angles $\vq = (q_1, q_2)$ and the disk position~$\vc$.
The finger angles follow from the tangency of each finger to the disk. Finger~$i$ lies on the side~$\nu_i$ of the symmetry axis, with $\nu_1 = 1$ and $\nu_2 = -1$, and its joint is at $\vp_i = (0, \nu_i d)$. Let $\vt_i$ be the unit vector along finger~$i$ and $\vn_i$ the unit vector normal to this finger, pointing toward the disk, as shown in Fig.~\ref{fig:model}. They are represented by
\begin{equation}
  \vt_{i} = \begin{pmatrix} \cos q_i \\ -\nu_i\sin q_i \end{pmatrix}, \qquad
  \vn_{i} = \begin{pmatrix} -\sin q_i \\ -\nu_i\cos q_i \end{pmatrix}.
  \label{eq:tn}
\end{equation}
Let $\va_i = \vp_i + \ell_i\vt_i$ be the contact point of finger~$i$, at distance~$\ell_i$ from its joint. Contact requires $\ell_i \le L$. At $\va_i$, the radius of the disk is normal to the finger, so that
\begin{equation}
  \vc - \vp_i = \ell_i\,\vt_i + r\,\vn_i .
  \label{eq:tangency}
\end{equation}
With reference to Fig.~\ref{fig:triangle},
let $h_{i} = d - \nu_i \cy$ be the distance along $y$ between joint~$i$ and the disk center. Since $\vt_i \perp \vn_i$, the points~$\vp_i$, $\va_i$, and $\vc$ form a right triangle with legs~$\ell_i$ and $r$. The segment from $\vp_i$ to $\vc$ is the hypotenuse of this triangle, and also of the right triangle with legs~$\cx$ and $h_i$. Therefore, $\ell_i^2 + r^2 = \cx^2 + h_i^2$. This segment makes the angle~$\arctan(h_i/\cx)$ with the $x$ direction and the angle~$\arctan(r/\ell_i)$ with the finger. Hence,
\begin{equation}
  \ell_i = \sqrt{\cx^2 + h_i^2 - r^2}, \qquad
  q_i = \arctan\frac{h_i}{\cx} - \arctan\frac{r}{\ell_i} .
  \label{eq:geometry}
\end{equation}

The velocities follow from the contact condition. The point of finger~$i$ at $\va_i$ moves with velocity $\ell_i\dot q_i\,\vn_i$, in the normal direction. The contact is rigid but frictionless: the normal velocities of the finger and the disk must match, and their tangential velocities can differ. Matching the normal velocities, $\ell_i\dot q_i = \vn_i^\top\dot{\vc}$, gives
\begin{equation}
  \dot q_i = \vg_i^\top \dot{\vc}, \qquad
  \vg_i = \frac{\vn_i}{\ell_i} =
  \nabla q_i ,
  \label{eq:velocity}
\end{equation}
where $\nabla q_i = \partial{q_i} / \partial \vc$ and the transmission vector $\vg_i = (\gcomp{i}{x}, \gcomp{i}{y})$ relates the velocity of the disk to that of finger~$i$. For both fingers, $\dot{\vq} = \mJ\,\dot{\vc}$, with the Jacobian $\mJ = [\vg_1\ \vg_2]^\top$.

\subsection{Dynamics}
\label{sec:dyn}
We derive the dynamics of the controlled gripper in the standard manipulator form. Let $f_i \ge 0$ be the magnitude of the force applied by finger~$i$ on the disk, and $f_0 \ge 0$ that of the palm, which acts along the palm normal $\vn_0 = (1, 0)$. Balancing forces and moments, we have
\begin{subequations}
\begin{align}
  m\,\ddot{\vc} &= \vF + f_1\vn_1 + f_2\vn_2 + f_0\,\vn_0 ,
  \label{eq:newton_disk}\\
  \If\,\ddot q_i &= \tau_i - \ell_i f_i , \qquad i \in \{1,2\} .
  \label{eq:newton_finger}
\end{align}
\end{subequations}
Solving \eqref{eq:newton_finger} for $f_i$ and substituting it into \eqref{eq:newton_disk} eliminates the finger forces. This step uses $\vn_i/\ell_i = \vg_i$ and $\ddot{\vq} = \mJ\,\ddot{\vc} + \dot{\mJ}\,\dot{\vc}$, the time derivative of $\dot{\vq} = \mJ\,\dot{\vc}$. The disk is then driven by the joint torques $\vtau = (\tau_1, \tau_2)$:
\begin{equation}
  \mM(\vc)\,\ddot{\vc}
  + \mC(\vc,\dot{\vc})\,\dot{\vc}
  = \mJ^\top\vtau + \vF + f_0\,\vn_0 ,
  \label{eq:openloop}
\end{equation}
where $\mM = m\,\mI_2 + \If\,\mJ^\top\mJ$ is the inertia matrix and $\mC = \If\,\mJ^\top\dot{\mJ}$ is the Coriolis and centripetal matrix. The term~$\If\,\mJ^\top\mJ$ is the inertia of the fingers as felt by the disk. The term~$\mJ^\top\vtau$ is the force of the joint torques on the disk: by power conservation, $\vtau^\top\dot{\vq} = (\mJ^\top\vtau)^\top\dot{\vc}$.

The PD controller \eqref{eq:pd} acts at each joint as a torsional damper of coefficient~$\kd$ in parallel with a torsional spring of stiffness~$\kp$ and rest angle~$\qref$.The potentials of the two springs sum to the \emph{controller potential}
\begin{equation}
  V(\vc) = \frac{\kp}{2} \sum_i e_i(\vc)^2 .
  \label{eq:potential}
\end{equation}
Since $\nabla e_i = -\vg_i$ by \eqref{eq:velocity}, the gradient of the controller potential is $\nabla V = -\kp\sum_i e_i\,\vg_i$, and $\dot e_i = -\vg_i^\top\dot{\vc}$. The joint torques therefore act on the disk as
\begin{equation*}
  \mJ^\top\vtau = \sum_i \vg_i\,\tau_i = \sum_i \vg_i\,(\kp e_i + \kd\dot e_i) = -\nabla V - \mD\,\dot{\vc},
\end{equation*}
with the damping matrix $\mD = \kd\,\mJ^\top\mJ$. The closed loop is therefore the mechanical system
\begin{equation}
  \mM(\vc)\,\ddot{\vc}
  + \mC(\vc,\dot{\vc})\,\dot{\vc}
  + \mD(\vc)\,\dot{\vc}
  + \nabla V(\vc)
  = \vF + f_0\,\vn_0 ,
  \label{eq:reduced}
\end{equation}
in the standard manipulator form: inertial, Coriolis, damping, and potential forces on the left, and the load and the palm force on the right. The control parameters enter only as constants, and the load~$\vF$ is the input. The controller sets the stiffness, through~$V$, and the damping~$\mD$. Both also depend on the position of the disk, through~$\mJ$.

The palm contact is unilateral, so the system has two modes. \emph{(i)}~\emph{Off the palm} ($\cx > r$), $f_0 = 0$, and \eqref{eq:reduced} is a two-degree-of-freedom system in $\vc$. \emph{(ii)}~\emph{On the palm} ($\cx = r$, $\dcx = \ddcx = 0$), the $x$ row of \eqref{eq:reduced} gives the palm force~$f_0 \ge 0$, and the $y$ row gives the lateral dynamics. The disk leaves the palm when this force would become negative. At the palm-release boundary ($\cx = r$, $f_0 = 0$), the disk touches the palm without pressing on it.

\sectionbreak
\section{Input-output characterization of a grasp}\label{sec:linearization}

\subsection{Admittance derivation via linearization}\label{sec:lin}

We characterize a grasp by its \emph{admittance}, the transfer matrix from load perturbations to displacements around an equilibrium.
A \emph{forced equilibrium} $(\cbar,\Fbar)$ is a position~$\cbar$ at which the disk rests under a constant load~$\Fbar$. There, the force that the controlled fingers apply to the disk balances the load and, on the palm, the palm force. At rest, \eqref{eq:reduced} reduces to
\begin{equation}
  \nabla V(\cbar) = \Fbar + f_0\,\vn_0 ,
  \label{eq:equilibrium}
\end{equation}
with $f_0 = 0$ off the palm ($\cxbar > r$) and $f_0 \ge 0$ on the palm ($\cxbar = r$). Off the palm, the equilibria under a load~$\Fbar$ are the stationary points of the \emph{total potential} $V(\vc) - \Fbar^\top\vc$.

Consider $\delta\vc = \vc - \cbar$ and $\delta\vF = \vF - \Fbar$. Since $\dot{\mJ}$ is linear in $\dot{\vc}$, the Coriolis term~$\mC\dot{\vc}$ is quadratic in $\dot{\vc}$ and has no first-order part. Linearizing \eqref{eq:reduced} therefore gives
\begin{equation}
  \mM\,\delta\ddot{\vc} + \mD\,\delta\dot{\vc}
  + \mK\,\delta\vc = \delta\vF,
  \qquad
  \mK = \nabla^2 V ,
  \label{eq:linearised}
\end{equation}
where $\mM$, $\mD$, and the stiffness matrix~$\mK$ are evaluated at $\cbar$. 
In the Laplace domain,
$\delta\vc(s) = \mH(s)\,\delta\vF(s)$, with 
\begin{equation}
  \mH(s) = \bigl(\mM s^2 + \mD s + \mK\bigr)^{-1} \in \mathbb{C}^{2\times 2} .
  \label{eq:H}
\end{equation}

The admittance \eqref{eq:H} is valid off the palm and at the palm-release boundary ($f_0 = 0$). On the palm ($f_0 > 0$), the palm force absorbs small perturbations of the axial load, and the disk does not move along $x$. The lateral admittance $1/(M_{yy}s^2 + D_{yy}s + K_{yy})$ still captures the lateral motion.

\subsection{Static metrics: stiffness and stability}\label{sec:stab}
At zero frequency, the admittance captures the \emph{stiffness} of the grasp, which also decides its \emph{stability}. Off the palm, a constant load perturbation~$\delta\vF$ causes the displacement~$\mH(0)\,\delta\vF$, where $\mH(0) = \mK^{-1}$.
The inertia matrix~$\mM$ is positive definite, and so is the damping matrix $\mD = \kd\,\mJ^\top\mJ$ when the two finger normals are not parallel. Therefore, off the palm, all poles of $\mH$ have a negative real part if and only if $\mK$ is positive definite~\cite{thomson_tait1879treatise, howard1996stability}. The grasp is unstable if $\mK$ has a negative eigenvalue. Stability is therefore read from the curvature of the potential at the equilibrium. At the boundary of stability, $\mK$ is singular: the grasp has no stiffness in some direction, and a small load moves the disk by a large amount.  On the palm, away from the palm-release boundary, the grasp is stable if and only if $K_{yy} > 0$.

\subsection{Dynamic response and resonances}\label{sec:features}

At nonzero frequency, the admittance captures the response to time-varying load perturbations. The modulus of each of its  poles, for example, captures a \emph{natural frequency} of the grasp, near which load perturbations are amplified.  Likewise, the \emph{damping ratio} of each pole, the absolute value of the ratio of its real part to its modulus, captures the response oscillations. 

The maximal amplification that a time-varying load produces is classically captured by the \emph{peak gain} $\|\mH\|_\infty = \max_\omega \bar\sigma\bigl(\mH(j\omega)\bigr)$, where $\bar\sigma$ is the largest singular value. The maximum can also be restricted to task-relevant frequencies $\omega \in [0,\bar{\omega}]$.  Compared with the static value, or DC gain, $\bar\sigma\bigl(\mH(0)\bigr)$, the peak gain tells how much the stiffness underestimates the displacement under time-varying perturbations.

These considerations can be further specialized to the \emph{diagonal terms} of $\mH(s)$, which relate a load perturbation to the displacement in the same direction, and to its
\emph{off-diagonal terms}, which capture the displacement in one direction caused by a load perturbation in another direction, such as the lateral displacement caused by an axial load. 

Two grasps with the same stiffness matrix can differ in their dynamic response. The admittance captures these differences, and therefore the effects of accelerations and impacts in grasping. It also maps the gripper parameters directly to grasp features: each choice of these parameters gives a specific linearization, hence specific poles and peak gains. This gives a direct link between the gripper parameters on the one hand and the resonances, peak gain, and stability of the grasp on the other. Within the simplified setting of this paper, the admittance reduces to a diagonal matrix, which captures decoupled responses along the $x$ and $y$ axes.

\sectionbreak
\section{Static analysis of symmetric grasps}\label{sec:static}

For symmetric grasps, the equilibria and their stability are a function of the holding force and depend on two key quantities.  A grasp is symmetric when the load acts along the symmetry axis, $\vF = (F_x, 0)$ with $F_x \ge 0$, and the disk is centered, $\cy = 0$. Both fingers then have the same angle~$q$, error $e = \qref - q$, and contact distance~$\ell$. For such grasps, the axial and lateral motions are decoupled: the matrices of Section~\ref{sec:lin} are diagonal, $\mM = \operatorname{diag}(M_{xx},M_{yy})$, $\mD = \operatorname{diag}(D_{xx},D_{yy})$, and $\mK = \operatorname{diag}(K_{xx},K_{yy})$.

\subsection{Stability and bifurcation analysis}\label{sec:bif}

\subsubsection*{Lateral stability} The lateral stiffness is positive in every symmetric grasp. Since $\partial e_i/\partial \cy = -\gcomp{i}{y}$, it reads
\begin{equation}
    K_{yy} = \kp\sum_i\bigl(\gcomp{i}{y}^2 - e\,\partial \gcomp{i}{y}/\partial \cy\bigr).
    \label{eq:Ky}
\end{equation}
The first term is positive. For the second term, $\gcomp{i}{y} = -\nu_i\cos q_i/\ell_i$. When the disk moves toward finger~$i$, that is, in the direction~$\nu_i$, this finger opens and its contact point moves toward the joint, so $\cos q_i/\ell_i$ grows. Hence, $\partial \gcomp{i}{y}/\partial \cy < 0$ for both fingers. Finally, $e \ge 0$ at rest, because $\kp e = \ell f_i$ and the fingers can only push. Hence, $K_{yy} > 0$. On the palm, a symmetric grasp is therefore always stable. Off the palm, its stability depends on the axial stiffness~$K_{xx}$.

\subsubsection*{Holding force and equilibria} Along the axis, the equilibria are the positions where the \emph{holding force}, the force with which the fingers push the disk toward the palm, balances the load. By \eqref{eq:geometry}, $q$ and $\ell$ depend only on~$\cx$, so the holding force is a function of $\cx$ only. The disk moves from the palm, $\cx = r$, to the fingertip, $\cx = \xtip$, where $\ell(\xtip) = L$. Recall that $d > r$, so $q(\cx) > 0$.

Since $\cy = 0$, the $y$-components of $\vg_1$ and $\vg_2$ cancel out. The fingers push the disk toward the palm with the force $-\nabla V = -P(\cx)\,\vn_0$, where 
\begin{equation}
  P(\cx) = 2\kp\,e(\cx)\,\sigma(\cx), \quad
  \sigma(\cx) = -q'(\cx) = \frac{\sin q(\cx)}{\ell(\cx)} ,
  \label{eq:pull}
\end{equation}
where $e(\cx) = \qref - q(\cx)$ is the error and the prime denotes the derivative with respect to $\cx$. 
The \emph{transmission} $\sigma$ tells how fast the fingers open when the disk moves away from the palm: a displacement~$\delta\cx$ opens each finger by $\sigma\,\delta\cx$. Its expression follows from $\vg_i = \nabla q_i$ in \eqref{eq:velocity}. By virtual work, the joint torque~$\kp e$ of each finger therefore pushes the disk toward the palm with the force~$\kp e\,\sigma$, and the two fingers together give $P$. The holding force is positive wherever the fingers press on the disk ($e > 0$).

The torque and the transmission vary in opposite directions. When the disk moves away from the palm, the fingers open, so $e$ and the torque grow. Meanwhile, the contact distance~$\ell(\cx)$ increases and the fingers become more aligned with $x$, which together reduce $\sigma$.

\begin{figure}
    \centering
    \input{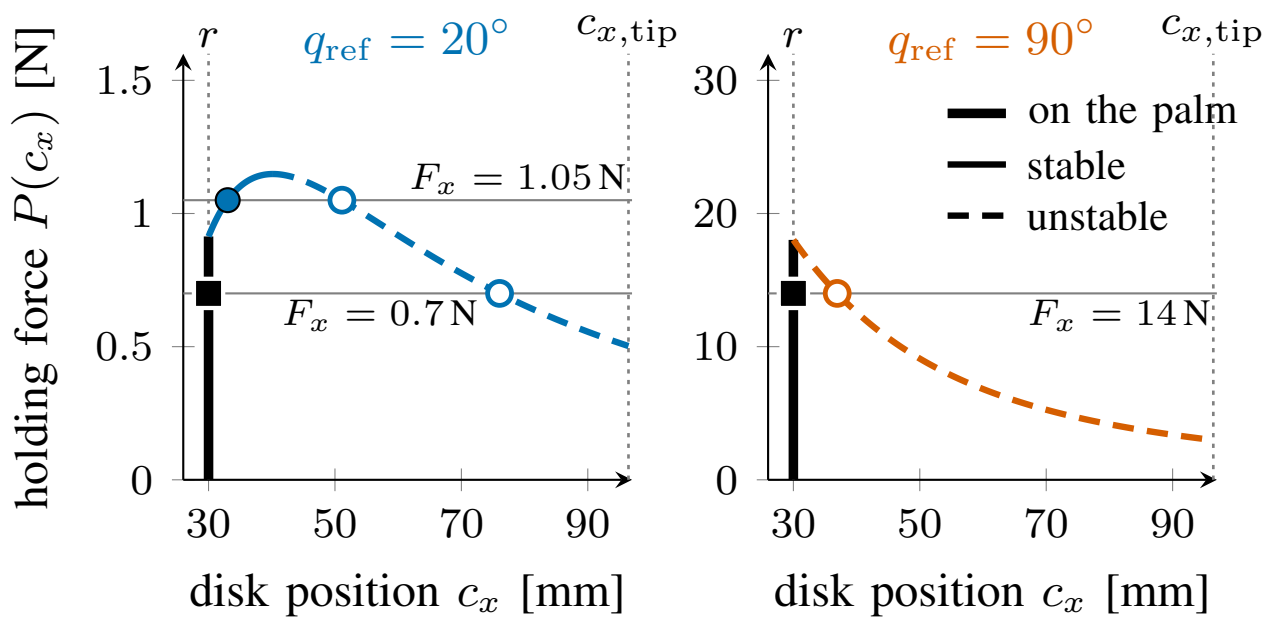}
    \caption{The reference sets the shape of the holding force~$P(\cx)$: with $\qref = \qty{20}{\degree}$ (left), $P$ rises then falls; with $\qref = \qty{90}{\degree}$ (right), it decreases from the palm. Both use the same gain~$\kp$, but the holding force on the right is much larger. The equilibria lie where the horizontal line~$F_x$ meets the curve: stable (filled dots, solid) and unstable (empty dots, dashed) off the palm, and on the palm (square) while $F_x \le P(r)$.}
    \label{fig:holding}
\end{figure}

The $x$ row of the equilibrium condition \eqref{eq:equilibrium} reduces to $f_0 = P(\cx) - F_x$, and its $y$ row holds by symmetry. The disk therefore rests on the palm, $\cx = r$, if and only if $P(r) \ge F_x$, with the palm force $f_0 = P(r) - F_x$. Off the palm, the disk is at equilibrium at every $\cx \in (r, \xtip]$ such that $P(\cx) = F_x$. Graphically, these equilibria are the intersections of the curve~$P(\cx)$ with the horizontal line~$F_x$. Their number and positions depend on the gripper parameters, in particular on the reference~$\qref$: depending on $\qref$, $P(\cx)$ either rises then falls or decreases from the palm (Fig.~\ref{fig:holding}).

\subsubsection*{Axial stability} Off the palm, an equilibrium is stable if and only if the holding force increases with the position~$\cx$ of the disk. Stability depends on the sign of the axial stiffness~$K_{xx}$. Since $\cy = 0$, $P = \partial V/\partial \cx$, so $K_{xx} = P'(\cx)$. The reference is constant, so the error grows as the fingers open: $e' = -q' = \sigma$. Differentiating \eqref{eq:pull} then gives
\begin{equation}
    \begin{aligned}
        K_{xx} &= 2\kp\,(\sigma^2 + e\,\sigma') \\
        & = \frac{2\kp\sin q}{\ell^2}\Bigl[\sin q - e\Bigl(\cos q + \frac{\cx}{\ell}\Bigr)\Bigr].
        \label{eq:Kx}
    \end{aligned}
\end{equation}
In the last expression, we have used $q' = -\sin q/\ell$ and $\ell' = \cx/\ell$. The first expression, $2\kp\,(\sigma^2 + e\,\sigma')$, combines the transmission~$\sigma$ with the transmission variation~$\sigma'$, weighted by the error~$e$. Their competition decides the sign of $P'(\cx)$: stable equilibria lie on the rising parts of $P(\cx)$, unstable ones on the falling parts (Fig.~\ref{fig:holding}). This split recovers the known decomposition of the grasp stiffness into the gain seen through the Jacobian and a geometric term proportional to the joint torques~\cite{cutkosky1989computing, chen2000conservative}. For this gripper, the geometric term is always negative, because the transmission decreases as the disk moves away from the palm.

The factor $\frac{2\kp\sin q}{\ell^2}$ in \eqref{eq:Kx} is positive, so the sign of $K_{xx}$ is that of the term in square brackets. An equilibrium off the palm at $\cx$ is therefore stable if
  \begin{equation}
    e < \estar(\cx) = \frac{\sin q(\cx)}{\cos q(\cx) + \cx/\ell(\cx)} ,
    \label{eq:estar}
  \end{equation}
and unstable if $e > \estar(\cx)$. Equivalently, $\estar = \sigma^2/|\sigma'|$: it is the critical error at which the relative growth of the joint torque, $\sigma/e$, equals the relative loss of transmission, $|\sigma'|/\sigma$.

\begin{figure}
    \centering
    \input{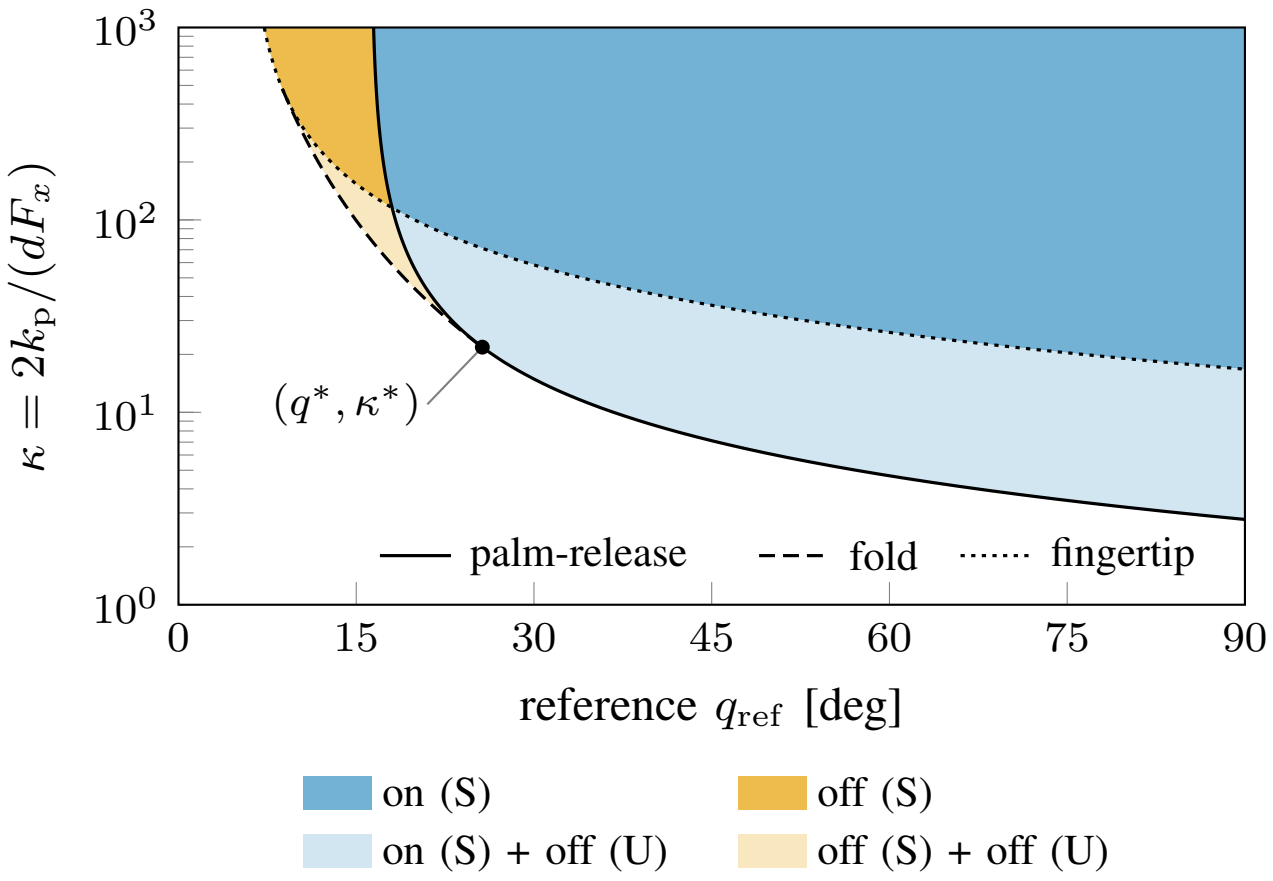}
    \caption{The reference~$\qref$ and the normalized gain~$\kappa$ set the number, location (on or off the palm), and stability (S: stable, U: unstable) of the axial equilibria, for all loads and gains. Three curves separate the regions: palm-release (solid), fingertip (dotted), and fold (dashed).}
    \label{fig:bifurcation}
\end{figure}

\subsubsection*{Bifurcation plane} For a given gripper, the equilibria and their stability depend on two key quantities: the reference~$\qref$ and the normalized gain $\kappa = 2\kp/(d\,F_x)$. In fact, the holding force \eqref{eq:pull} is proportional to $\kp$, so the equilibrium condition $P(\cx) = F_x$ reads
\begin{equation}
  \kappa\,d\,e(\cx)\,\sigma(\cx) = 1 ,
  \label{eq:scaled_equilibrium}
\end{equation}
where $\kappa$ and $d\,e\,\sigma$ are dimensionless, with $e$ in radians. The reference~$\qref$ sets the shape of the holding force, and the normalized gain~$\kappa$ sets its scale. Doubling both the load and the gain leaves every equilibrium unchanged. The plane~$(\qref, \kappa)$ therefore summarizes the equilibria and their stability for all loads and gains (Fig.~\ref{fig:bifurcation}): moving up in this plane means decreasing the load at a given gain or, equivalently, increasing the gain under a given load. 

The plane is divided into regions according to the number, location, and stability of the equilibria. At a fixed gain, moving up a vertical line in Fig.~\ref{fig:bifurcation} is equivalent to lowering
the horizontal line~$F_x$ in Fig.~\ref{fig:holding}, whose intersections with the curve~$P(\cx)$ determine the equilibria and their stability. 

Three curves separate the regions of Fig.~\ref{fig:bifurcation}. On the palm-release curve (solid), $F_x = P(r)$: the palm force vanishes and the disk leaves the palm. On the fingertip curve (dotted), $F_x = P(\xtip)$: an off-palm equilibrium reaches the fingertip. On the fold curve (dashed), $F_x$ equals the maximum of $P$: the stable and unstable off-palm equilibria merge and disappear in a saddle-node bifurcation. Beyond this curve, the load exceeds the largest holding force, and the grasp is lost. The difference $\max_{\cx} P(\cx) - F_x$ is the axial \emph{static margin}, the largest additional constant load along $x$ for which an equilibrium remains: in Fig.~\ref{fig:holding}, the height of the top of the curve~$P$ above the line~$F_x$. Unlike the admittance, which is local, the static margin is global.

These curves delimit four regions with different equilibria. The dark regions have a single stable equilibrium, on the palm (blue) or off the palm (orange), separated by the palm-release curve. Crossing the fingertip curve toward the light regions adds an unstable equilibrium. It is the empty dot of Fig.~\ref{fig:holding}, and it bounds the basin of attraction of the stable equilibrium: a disk displacement beyond this point cannot be recovered. Elsewhere, no equilibrium exists and the grasp is lost. The fold and palm-release curves meet at $(q^\ast, \kappa^\ast)$. Stable equilibria off the palm require  $\qref < q^\ast$ and $\kappa > \kappa^\ast$.

\subsection{Effect of the load and the gripper parameters}\label{sec:obs}

\subsubsection*{Increasing the load}

For a fixed controller~$(\kp, \qref)$, slowly increasing the load~$F_x$ is equivalent to moving down along a vertical line in Fig.~\ref{fig:bifurcation}. We observe three qualitative behaviors. For large references ($\qref > q^\ast$), the vertical line crosses only blue regions: the disk stays on the palm until the load reaches the holding force at the palm, then leaves the palm, and the grasp is lost at once. For intermediate references, the vertical line crosses blue then orange regions: the disk rests on the palm, then is released from it and held by the fingers  until, for an even larger load, the grasp is lost. For small references, the vertical line crosses only orange regions: the disk never rests on the palm. Overall, the reference~$\qref$ sets the qualitative behavior, while the gain~$\kp$ scales the loads at which transitions occur. Large references and large gains therefore give a firmer grasp: they increase the static margin, so that an equilibrium remains under larger loads. This separation of roles can inform the choice of the control parameters for a given object.

\subsubsection*{Same equilibrium, different stiffness}

\begin{figure}
    \centering
    \input{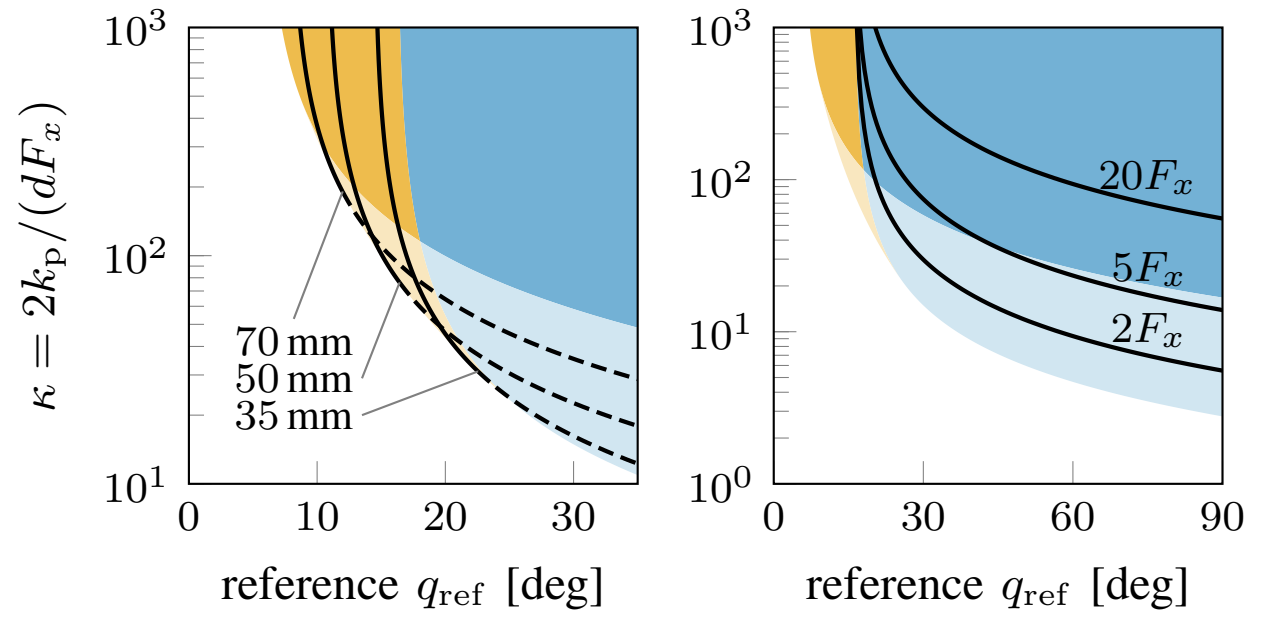}
    \caption{Along each black curve, the controllers~$(\kp, \qref)$ hold the disk at the same equilibrium. Moving up a curve raises $\kp$, lowers $\qref$ toward the finger angle, and raises the stiffness. \emph{Left:} curves of constant position, along which the disk rests off the palm at $\cx = \qtylist[list-units = single]{35;50;70}{\milli\metre}$. The equilibrium is stable on the solid part and becomes unstable on the dashed part, the two parts meeting on the fold. \emph{Right:} curves of constant holding force, $P(r) = 2F_x$, $5F_x$, and $20F_x$.}
    \label{fig:iso}
\end{figure}

Under a given load, different controllers~$(\kp, \qref)$ can hold the disk at the same equilibrium, that is, at the same position and with the same contact forces. At a given position~$\cx$, the force~$\kp e/\ell$ of each finger is unchanged if $\kp$ increases while the static error $e = \qref - q$ decreases in proportion. Each equilibrium thus corresponds to a family of controllers, represented by the black curves of Fig.~\ref{fig:iso}. 

The stiffness nevertheless changes: in $K_{xx}$ \eqref{eq:Kx} and $K_{yy}$ \eqref{eq:Ky}, the term proportional to $\kp e$ is constant along the curves of Fig.~\ref{fig:iso}, while the term proportional to $\kp$ is not. As $\kp$ decreases, $K_{xx}$ can even become negative, making the equilibrium unstable (dashed lines). From a design perspective, the equilibrium provides only partial information about the grasp. The controller~$(\kp, \qref)$ can be tuned to adjust the stiffness without disturbing the target equilibrium.

\begin{figure}
    \centering
    \input{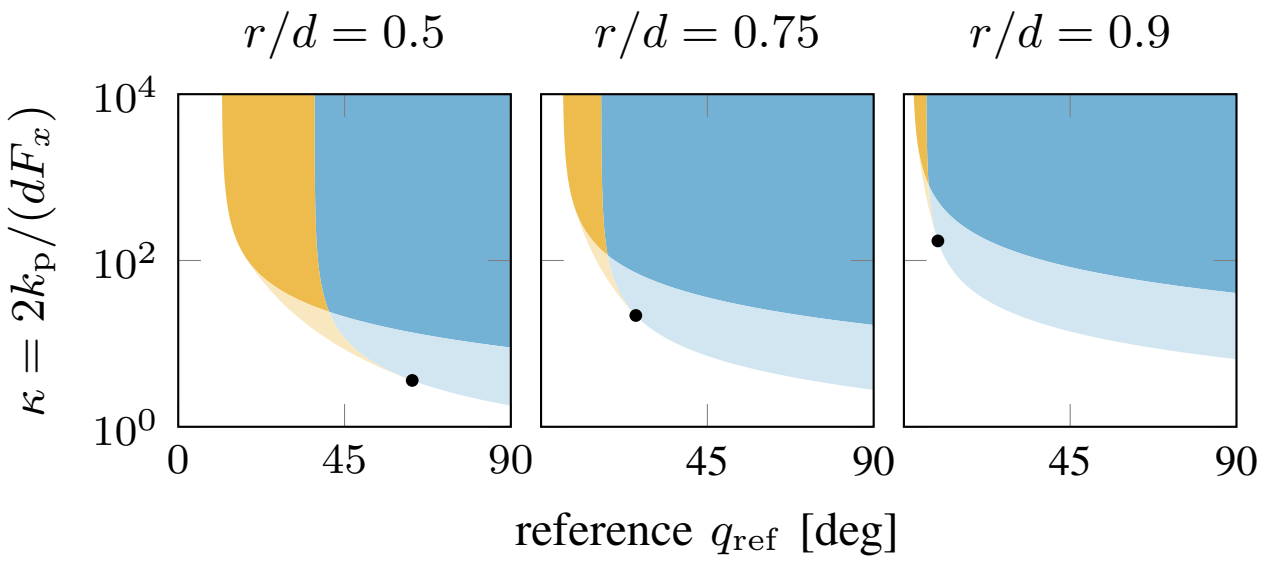}
    \caption{A larger disk moves the stable equilibria off the palm toward small references and large normalized gains. Bifurcation planes for $r/d = 0.5$, $0.75$, and $0.9$; the regions are those of Fig.~\ref{fig:bifurcation}, and the dots mark $(q^\ast, \kappa^\ast)$.}
    \label{fig:morphology}
\end{figure}

\subsubsection*{Changing the gripper morphology}

The ratio~$r/d$ of the disk radius to the half-distance between the finger joints affects the bifurcation plane, as shown in Fig.~\ref{fig:morphology}. For a large disk, with $r/d$ close to $1$, the fingers are nearly parallel to the $x$ direction and therefore transmit little of their torque along $x$. For example, from $r/d = 0.5$ to $0.9$, the critical $q^\ast$ drops from \qty{63}{\degree} to \qty{9}{\degree}, and $\kappa^\ast$ rises from $3.6$ to $172$. Holding a large disk off the palm therefore requires a small static error~$e$ and large normalized gains. In practice, only the equilibria on the palm can handle constant load perturbations.

\sectionbreak
\section{Dynamic responses of symmetric grasps}\label{sec:dynamic}

In what follows, we show how grasps that are equivalent under static metrics can exhibit different responses to time-varying load perturbations. We also show that the finger inertia is irrelevant from a static perspective but has a clear impact on the dynamic response. For illustration purposes, the derivative gain~$\kd$ is kept at its nominal value (Table~\ref{tab:symbols}). We support our discussion by varying the proportional gain~$\kp$ and the reference~$\qref$. Proper controller tuning, in which $\kp$ and $\kd$ are usually chosen together, is outside the scope of this paper.

Recall that the diagonal matrices of Section~\ref{sec:static} split the admittance into an axial and a lateral \emph{channel}, $H_{xx}(s) = 1/(M_{xx}s^2 + D_{xx}s + K_{xx})$ and $H_{yy}(s) = 1/(M_{yy}s^2 + D_{yy}s + K_{yy})$, each a mass-spring-damper system. Their stiffnesses are given by \eqref{eq:Kx} and \eqref{eq:Ky}, while their inertias and dampings are 
\begin{equation}
  \begin{aligned}
    M_{xx} &= m + 2\If\sigma^2, &
    D_{xx} &= 2\kd\sigma^2, \\
    M_{yy} &= m + 2\If\frac{\cos^2 q}{\ell^2}, &
    D_{yy} &= 2\kd\frac{\cos^2 q}{\ell^2}.
  \end{aligned}
  \label{eq:channel_coefficients}
\end{equation}
As they transmit their torque in the holding force \eqref{eq:pull},
the fingers transmit their inertia and their damping to the disk along $x$ through the transmission~$\sigma$.

\subsection{Same forced equilibrium, different dynamic responses}\label{sec:samepos}

Grasps that share the same forced equilibrium and have the same inertia and damping can differ in their stiffness, which alone changes their dynamic response. Along the thick black curve in Fig.~\ref{fig:isopos_bode} (left), the position of the disk and the contact forces are fixed. The transmission~$\sigma$ is then fixed, and so are $M_{xx}$ and $D_{xx}$, by \eqref{eq:channel_coefficients}. Because the equilibrium is fixed, the torque  $\kp e$ is also constant. Thus, the stiffness~$K_{xx}$ in \eqref{eq:Kx} depends only on $\kp$ and increases linearly with it. 

Moving up along the curve, toward higher gains~$\kp$ and smaller references~$\qref$, changes the dynamic response in three ways. First, the DC gain $H_{xx}(0) = 1/K_{xx}$ decreases, so  a constant load perturbation moves the disk less. Second, the damping ratio decreases, so the ratio of the peak gain to the DC gain, $\sup_\omega |H_{xx}(j\omega)| / H_{xx}(0)$, increases, as shown in Fig.~\ref{fig:isopos_bode} (right). For low proportional gains, the grasp thus responds almost uniformly over frequency and oscillates little in response to a step load perturbation. In contrast, for high proportional gains, the grasp has a clear resonance and oscillates longer in response to the same perturbation. Finally, the natural frequency of the grasp increases as the proportional gain increases. Therefore, only faster load perturbations are amplified.  However, an extremely high proportional gain does not necessarily make the grasp better, due to practical limitations and noise sensitivity. The lateral channel follows a similar trend.

\begin{figure}[htbp]
    \centering
    \input{fig/same_position_features_tikz}
    \caption{Grasps at the same forced equilibrium differ in their dynamic response: a higher $\kp$ with a lower $\qref$ raises the stiffness and the natural frequency and sharpens the peak. 
    \emph{Left:} three grasps on the curve $\cx = \qty{35}{\milli\metre}$ of Fig.~\ref{fig:iso} (left), at $(\qref, \kappa) = (\qty{19.5}{\degree}, 50)$~(\textit{blue}), $(\qty{17.0}{\degree}, 100)$~(\textit{rose}), and $(\qty{14.7}{\degree}, 1000)$~(\textit{gray}). 
    \emph{Right:} axial (top) and lateral (bottom) admittances of these grasps. Dashed vertical lines mark the natural frequency~$\omega_\mathrm{n}$, and shaded bars the ratio $\sup_\omega |H_{xx}(j\omega)| / H_{xx}(0)$. 
    }
    \label{fig:isopos_bode}
\end{figure}

\subsection{Same stiffness, different dynamic responses}\label{sec:samecomp}

Grasps with the same stiffness can still differ dynamically, because the damping transmitted by the fingers depends on the position of the disk. The thick curve in Fig.~\ref{fig:iso_Kx} (left) represents grasps with the same stiffness~$K_{xx}$. Unlike in the previous section, the disk  rests at different positions~$\cx$.

For the nominal parameters of Table~\ref{tab:symbols}, the fingers add  a small term~$2\If\sigma^2$ to the disk mass~$m$ in the inertia~$M_{xx}$ of \eqref{eq:channel_coefficients}. Both $M_{xx}$ and the natural frequency therefore barely change along the curve. The damping $D_{xx} = 2\kd\sigma^2$, however, decreases as the disk moves away from the palm, because the transmission~$\sigma$ decreases (Section~\ref{sec:bif}). The damping ratio therefore decreases, and the peak gain $\sup_\omega |H_{xx}(j\omega)|$ increases, as shown in Fig.~\ref{fig:iso_Kx} (right). Damping in the object, from its interaction with the environment, or friction at the contacts would reduce this effect, but not remove it.
These grasps also differ in their lateral stiffness (not shown).

A target axial stiffness therefore does not fully determine the dynamic response. 
Within the family of controllers that share the same stiffness, those that hold the disk closer to the palm, with a small proportional gain~$\kp$ and a high reference~$\qref$, achieve higher damping. 
Thus, they oscillate less in response to a step load perturbation and have a lower peak gain.

\begin{figure}[tbp]
    \centering
    \input{fig/iso_stiffness_xx_tikz}
    \caption{Grasps with the same stiffness differ in their peak gain, which grows as the disk moves away from the palm. \emph{Left:} curves of constant stiffness $K_{xx} = \qtylist[list-units = single]{20;80;320}{\newton\per\metre}$, for the load $F_x = \qty{0.6}{\newton}$. Grasps A, B, and C lie on the \qty{20}{\newton\per\metre} curve, at $\cx = \qty{60}{\milli\metre}$ (A, \emph{green}), $\qty{40}{\milli\metre}$ (B, \emph{orange}), and $\qty{32}{\milli\metre}$ (C, \emph{black}). 
    \emph{Right:} their axial admittance~$|H_{xx}|$, with a linear zoom near~$\omega_\mathrm{n}$; the DC gain~$H_{xx}(0)$ is the same.}
    \label{fig:iso_Kx}
\end{figure}

\subsection{The role of finger inertia}\label{sec:inertia}

The finger inertia affects neither the equilibrium nor the stiffness of a grasp, but it changes its dynamic response. Static metrics are independent of finger inertia and therefore miss this effect. By \eqref{eq:channel_coefficients}, the finger inertia~$\If$ enters only $M_{xx}$ and $M_{yy}$. A larger finger inertia therefore lowers the natural frequency and the damping ratio of each channel and raises its peak gain.

\begin{figure}[tbp]
    \centering
    \input{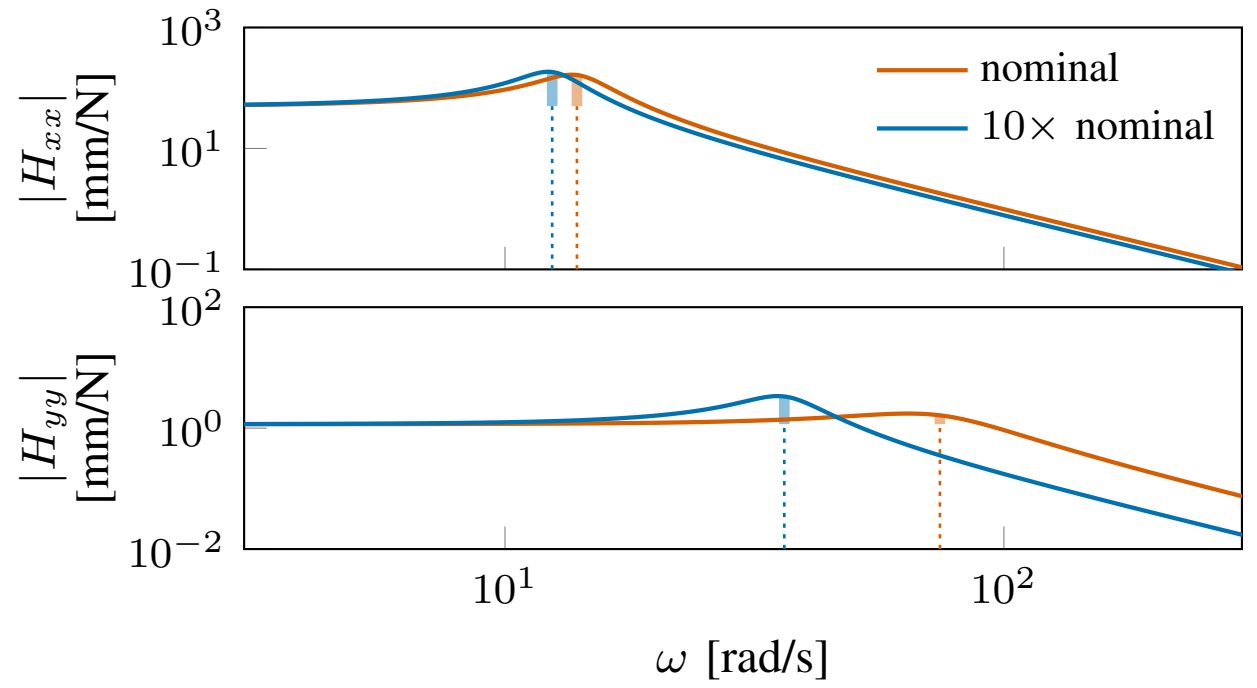}
    \caption{A larger finger inertia lowers the natural frequency, mostly in the lateral channel, where a peak appears. Axial (top) and lateral (bottom) admittances of grasp B of Fig.~\ref{fig:iso_Kx}, for the nominal finger inertia $\If = \qty{6.7e-5}{\kilo\gram\metre\squared}$ and an inertia 10 times larger. The stiffness is the same.}
    \label{fig:finger_inertia}
\end{figure}

The two channels are not affected equally. By \eqref{eq:channel_coefficients}, the finger inertia adds $2\If\sin^2 q/\ell^2$ to $M_{xx}$ and $2\If\cos^2 q/\ell^2$ to $M_{yy}$, in the ratio $\cot^2 q$. It therefore affects the lateral channel more when the fingers are closer to the $x$ direction ($q < \qty{45}{\degree}$), and the axial channel more otherwise. Which case occurs depends only on the geometry: $q$ is largest at the palm, where it exceeds \qty{45}{\degree} only for a small disk, with $r/d < 0.41$. With the nominal geometry, all grasps are in the first case. For grasp~B, a larger finger inertia lowers the lateral damping ratio enough for a resonance peak to appear, whereas the axial response barely changes (Fig.~\ref{fig:finger_inertia}).

The controller only partly compensates for the finger inertia. A higher $\kd$ restores the damping ratio, hence lowers the peak gain, but not the natural frequency. A higher $\kp$ restores the natural frequency only by raising the stiffness, which changes the response to constant loads and further lowers the damping ratio. Lighter fingers avoid this trade-off.

\subsection{Nonlinear responses}\label{sec:nonlinear}

\begin{figure}[tbp]
    \centering
    \input{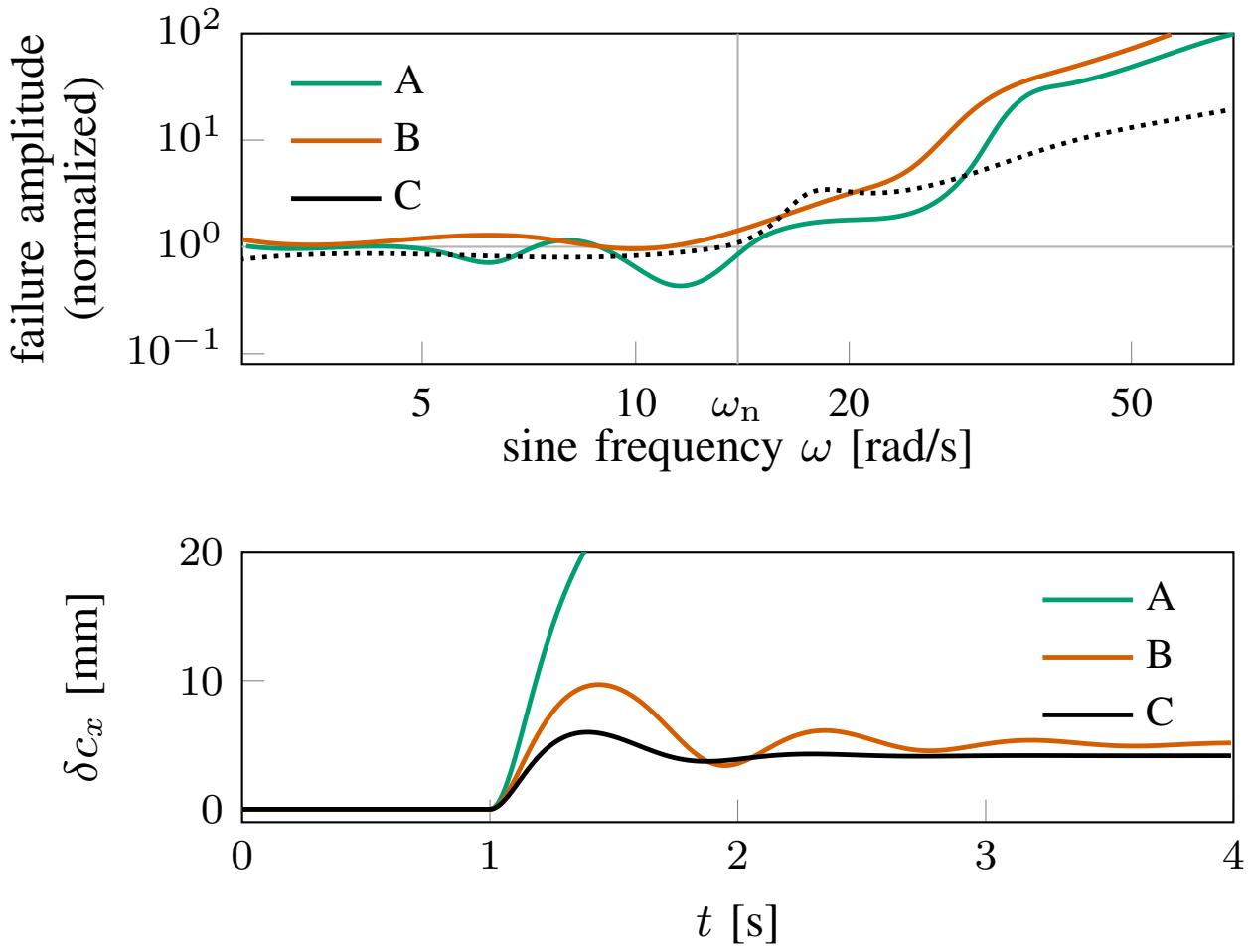}
    \caption{Nonlinear simulations of grasps A, B, and C of Fig.~\ref{fig:iso_Kx}. The failure amplitude of the nonlinear model is correlated with the axial admittance. \emph{Top:} smallest amplitude of a sinusoidal axial load perturbation that ejects the disk (solid) or makes it land on the palm (dotted), normalized by the static margin (horizontal line). The vertical line marks the natural frequency~$\omega_\mathrm{n}$ of the linearized model, shared by the three grasps. \emph{Bottom:} response to an axial step of 0.85 times the static margin: A is ejected, B and C settle. 
    }
    \label{fig:nonlinear}
\end{figure}

In this section, we show that the small-signal admittance captures relevant features of the full nonlinear behavior. We simulated the full dynamics \eqref{eq:reduced} for grasps A, B, and C of Fig.~\ref{fig:iso_Kx}, under sinusoidal axial load perturbations of fixed frequency. The amplitude of the load perturbation is slowly increased, which moves the grasp progressively from the small-signal regime to the nonlinear regime. For each frequency, we recorded  the smallest amplitude that made the grasp fail (Fig.~\ref{fig:nonlinear}, top). A grasp fails when the disk is ejected or when it hits the palm. The impact with the palm is a second failure mode, which the admittance does not describe, since it assumes that the disk moves freely between the fingers. Because an impact already counts as a failure, each simulation stops at the first impact. The amplitude is normalized by the static margin of Section~\ref{sec:bif}.

The failure curves are well correlated with the axial admittance. Below $\omega_\mathrm{n}$, the grasps fail at the static margin, as under a constant load. However, near $\omega_\mathrm{n}$, grasp A fails at an amplitude that is about $0.45$ times the static margin. This is expected, given the high peak gain of A in Fig.~\ref{fig:iso_Kx}. In contrast, grasp B fails close to the static margin up to $\omega_\mathrm{n}$.  For both, above $\omega_\mathrm{n}$, the inertia filters the load perturbation, and the failure amplitude grows with frequency. Grasp C, closer to the palm, lands on it before being ejected (dotted). The palm then blocks the axial motion and the dynamics change abruptly. This scenario lies outside the domain of validity of the linearized model.

The small-signal analysis, however, does not predict the large variations of the curve of A compared with B (Fig.~\ref{fig:nonlinear}, top). As the disk moves along the fingers, the stiffness of the grasp changes. This nonlinearity distorts the response and spreads the effect of the resonance over a wider range of frequencies. 

The admittance also predicts the axial step response well (Fig.~\ref{fig:nonlinear}, bottom). A step of $0.85$ times the static margin of each grasp ejects A, whereas B and C settle, with larger oscillations for B than for C. A step excites all frequencies, so its overshoot grows with the peak gain, which is highest for A and lowest for C (Fig.~\ref{fig:iso_Kx}). The overshoot of A carries the disk out of the fingers, although the step stays \qty{15}{\percent} below its static margin.

\section{Conclusion}\label{sec:conclusion}
Using a minimal gripper, we showed that neither the equilibrium nor the stiffness of a grasp fully determines its response to load perturbations. The admittance captures this response, demonstrating quantitatively that statically equivalent grasps can exhibit distinct behaviors when interacting with time-varying external perturbations. These grasps differ through the gain and the reference, the position of the disk, and the finger inertia. Nonlinear simulations confirmed the main features predicted by the admittance.

Our simplified gripper allows analytical derivations, but the admittance can also be obtained for general 
grippers, either numerically or experimentally. This matters for real grippers, which have nonlinear damping and static friction, and whose rotor inertia, multiplied by the square of the gear ratio, can dominate the finger inertia~\cite{bhatia2019direct}. Furthermore, real objects are rarely symmetric, and measurement noise, the sampling of a digital controller, and actuator saturation limit the gains that can be used.

We will study these effects in future work. This study was motivated by our ongoing research on anthropomorphic robotic hands~\cite{vanderheyden2026grasping}, where our control approach extends the PD control studied here to a network of virtual springs and dampers. Our next step is to compute the admittance of such a hand by automatic differentiation of its analytical model, and to validate it experimentally.

\addtolength{\textheight}{-12cm}   








\section*{Acknowledgments}
The authors used Claude (Anthropic) to help implement the code of Sections~\ref{sec:static} and~\ref{sec:dynamic} and the TikZ code of the figures.
The authors reviewed and edited the content as needed and take full responsibility for the publication.

\balance
\bibliographystyle{IEEEtran}
\bibliography{bibliography_corrected}

\end{document}